\documentclass[a4paper]{article}
\usepackage[utf8]{inputenc}
\usepackage[T1]{fontenc}
\usepackage[english]{babel}
\usepackage{textcomp}
\usepackage{mathtools,amssymb,amsthm}
\usepackage{lmodern}
\usepackage{geometry}
\usepackage{graphicx}
\usepackage{xcolor}
\usepackage{array}
\usepackage{calc}
\usepackage{titlesec}
\usepackage{titletoc}
\usepackage{fancyhdr}
\usepackage{titling}
\usepackage{enumitem}
\usepackage[hidelinks]{hyperref}
\hypersetup{pdfstartview=XYZ}
\usepackage{authblk}
\usepackage{multicol}
\usepackage{float}
\usepackage[hang,small,labelfont=bf,up,textfont=it,up]{caption}
\title{New methods for SVM feature selection}
\author{Tangui Aladjidi, François Pasqualini}
\affil{École polytechnique - ST Microelectronics}

\date{August 2018}
\usepackage{tocbibind}
\usepackage{subfigure}
\begin{document}
\begin{figure}[t]
\subfigure{\includegraphics[scale=0.5]{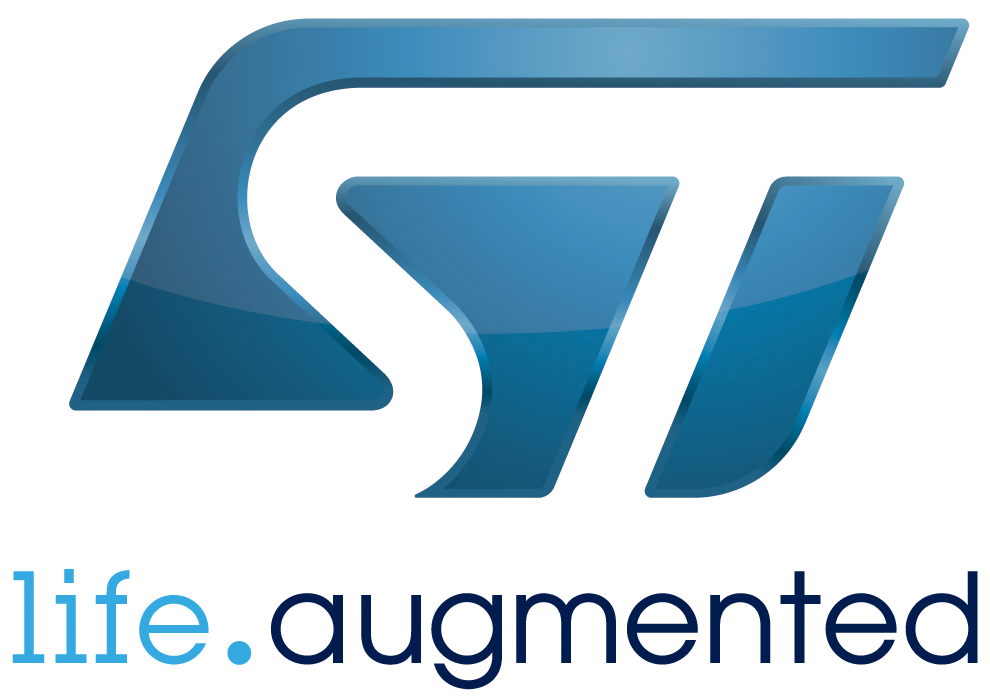}}
\hfill
\hspace*{-3cm}
\subfigure{\includegraphics[scale=0.1]{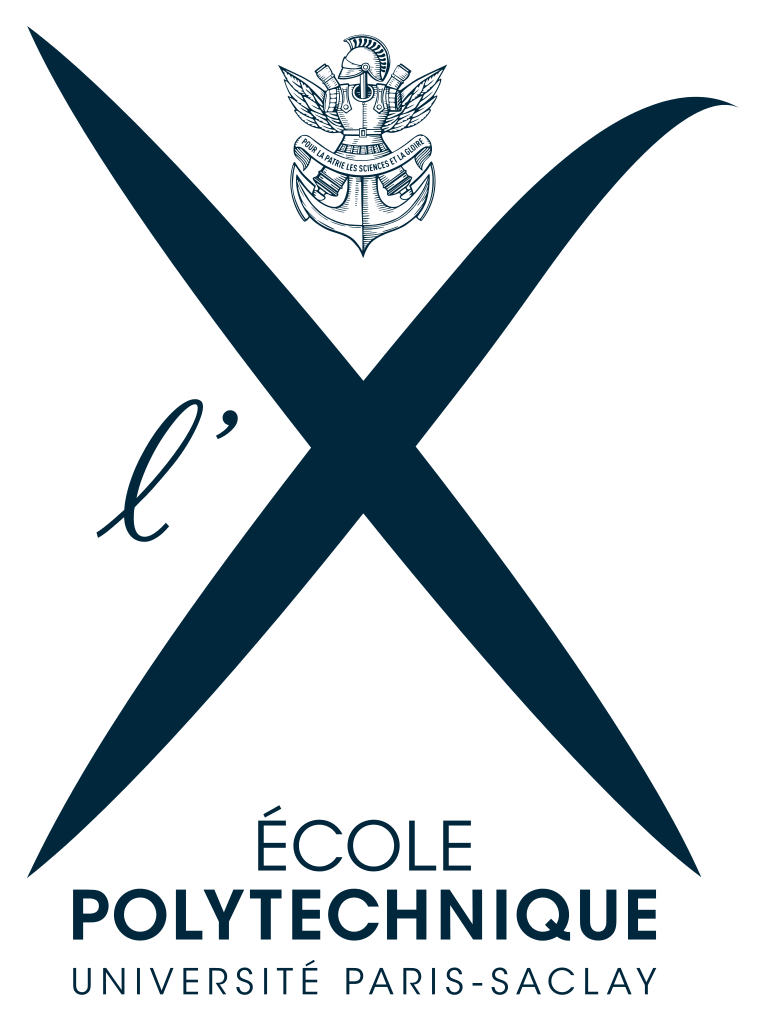}}
\hfill
\end{figure}
\maketitle
\section{Introduction}
\label{intro}
Support Vector Machines have been a popular topic for quite some time now, and as they develop, a need for new methods of feature selection arises. This work presents various approaches SVM feature selection developped during the author's summer internship at STMicroelectronics. The work focuses on the use of one-class SVM's for wafer testing. A key problem in OC-SVM algorithms is dimensionality reduction, otherwise known as feature selection. Prior to the execution of the proper SVM part of the algorithm, the program first needs to asses what will be the most significant data to take into account. Until now, approaches presented in \cite{ali} relied either on standard deviation estimators, or on recursive feature selection \cite{Huang}. This paper presents a new computationally efficient approach using Shannon's entropy as an information estimator to tackle pathological data sets, as well as a combinations methods of various already developped feature selection to drastically improve model accuracy and computation time. Finally, improvements to RFE techniques using K-Medoid clustering are also presented. All of these methods were tested on wafer fabrication data provided by STMicroelectronics assembly lines in Crolles (France), which allowed for good benchmarking as it makes for a very sizeable labeled data base.
The data used for developpement and testing is confidential, so we will present anonymous data throughout the whole paper. The numerical simulations were carried out using R.

\begin{multicols}{2}

\section{Data structure and general context}
\label{sec:1}
This work is based on wafer testing data, that is very large matrices where a column corresponds to a test parameter (throughout the paper, we will often refer to these), and a line corresponds to the tested lot. As we have no prior knowledge on the data, that is whether the wafer is good or bad, we can only use one-class classification : OC-SVM.
This numerical data is a challenge in two main aspects :
\begin{itemize}
	\item{the sheer size of the data calls for computationally efficient feature selection methods}
	\item{The mathematical foundations of SVM's require to work on the restriction to a more meaningful subspace. This is particularly true on "sparse" data}
The goal of this work was to obtain the best ratio between false detection rate and detection rate, that is, detecting all defective lots and only these lots. The goal was also to develop methods that would work with the two main types of test run on the wafers : parametric and yield tests. Parametric tests measure values of specific components on the wafers, and thus take continuous values in a given interval. Yield tests on the contrary, measure error rates at the very end of the production process, and thus generate very sparse and/or constant data as a lot of the parameters have a very low error rate. This type of "pathologic" data was in fact key to making efficient variable selection methods. \par
The more specific challenge of variable selection is then to choose the most meaningful parameters, and to try and favor complementary information, rather than redundant information on the wafer's distribution.
\end{itemize} 

\section{Univariate approaches : MADe and entropy criteria}
\label{sec:2}
The work started by trying to improve on \cite{ali}
's work, in which a very simple yet robust approach was used for feature selection : the Maximum Absolute Deviation estimator. For a set of observations $\left\lbrace x_i \right\rbrace $ of a variable $x$, it is given by the following formula :
\begin{equation}
	m=med(|x_i-med(\left\lbrace x_i \right\rbrace)|)
\end{equation}
This is a standard deviation estimator whose tolerance to abnormal values is very high, and thus was very conveniant for our use case. From this estimator we can then easily derive a confidence interval, and then a percentage of out of distribution points (OOD). This percentage is defined as follows :
\begin{equation}
	OOD_{MADe}=\frac{1}{Card\left\lbrace x_i \right\rbrace}Card\left\lbrace | x_i | \geq n\times 1,483 \times m \right\rbrace
\end{equation}
The 1.483 factor is an empirical factor derived from the convergence of this estimator to the standard deviation. The factor n allows us to tune the width of the set.\par
However, it quickly shows its limits on simple but pathological sets of data. Indeed, this estimator tends to select sets that have a few out of distribution points rather than more "tame" distributions that carry more information. For practical purposes this is shown particularly on sparse sets of data, that is yield data that contains a lot of zeros or constant variables. 
\subsection{Entropy measure and criterion.}
To overcome this tendancy, we defined a new measure of a distribution's entropy, based on Shannon's entropy. This measure is constructed very easily by computing a histogram of a data set $\lbrace x_i \rbrace$. We can then associate to each $x_i$ a empirical probability. We set the outer bins of the histogram to be $-\infty$ and $+\infty$ so that we can use one binning for several data sets (this is capital for the actual main machine learning algorithm). From this empirical probability distribution we can then derive the distribution's entropy :
\begin{equation}
	H=-\sum_{i} p_i log(p_i)
\end{equation} 
Where $p_i$  is the empirical probability associated to each variable sample $x_i$. We then sort the variables according to their entropy : the more the better. One main advantage of this measure is that it has a much higher tolerance to aberrant values than MADe. This was tested by artificially introducing arbitrarily high values in the data sets, and checking if the ranking was preserved.\par
\begin{figure}[H]
	\includegraphics[scale=0.52]{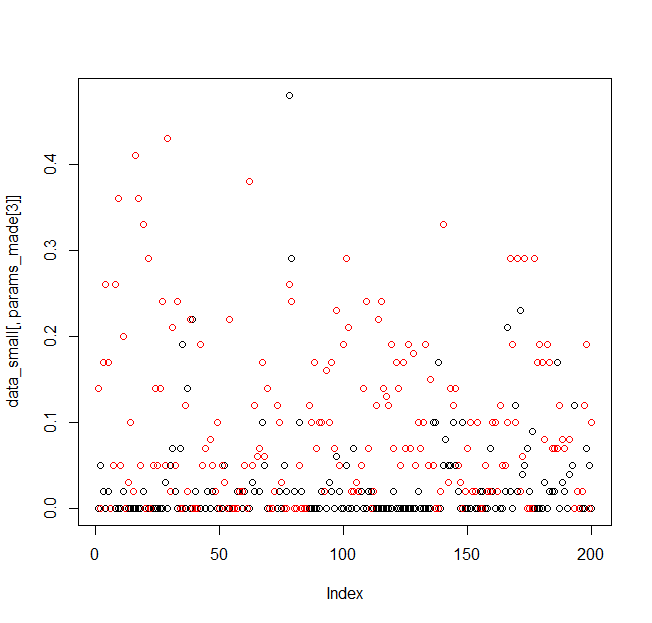}
	\caption{The two first variables according to MADe (black) and entropy criteria (red).}
	\label{made_vs_entropy}
\end{figure}

We compare these two methods by comparing the best ranked variables on various sparse data sets, which are the most discriminant for our use case, as seen in fig \ref{made_vs_entropy}.

We can see in figure \ref{made_vs_entropy} that MADe attributes a good ranking to variables having a main baseline of zeroes with little to no recognizable pattern. On the contrary, the variable having the same rank but using an entropy ranking has a lot more information in it. 

\section{Multivariate approaches : recursive feature elimination.}
\subsection{RFE algorithm presentation.}
The previous approaches are efficient, but they only focus on a single variable at a time. However, it is known that for SVM's, multivariate approaches are to be preferred as they tend to produce more robust models. The final goal is indeed to have a variable selection method, that is really embedded in the SVM scheme by taking advantage of its predictive abilities. \par
We first implemented the now classical approach of Recursive Feature Selection (RFE). This method is presented in more details in \cite{Huang}. Basically, the principle of RFE is to recursively eliminate variables that have the least influence in the model. Here is a succint presentation of the algorithm :
\begin{itemize}
\item{Step 0 : We initiate a set of $n$ variables $\mathcal{S}=\lbrace (x_{i}^{(k)})_{i\in [0,N]} | k \in [0,n] \rbrace$ ($N$ is the number of observations of each variable).}
\item{Step 1 : We build a one-class SVM model $\mathcal{M}$ of $\mathcal{S}$.}
\item{Step 2 : We sort each variable according to $|w_i|^2$, where $w_i$ is the support vector corresponding to each variable $x^{(k)} \in \mathbb{R}^N$.}
\item{Step 3 : Let $f$ be the index such that $f=\underset{i\in [0,N]}{argmin}(|w_i|^2)$. We now remove the feature $x^{(f)}$ with the least such that $\mathcal{S}=\mathcal{S}\setminus x^{(f)}$.}
\item{Repeat step 1 through 3 until $\mathcal{S}$ reaches target size.}
\end{itemize}
Target size was defined through a grid search with the parameters of the OC-SVM on various sample sets, and is independent of the variable selection method.
This method is very computationaly inefficient, it can however be greatly optimized using parallel computing, and by removing more than a variable at a time.\par
\subsection{Classification and clustering properties.}
RFE's  main advantage is its exceptionnal performance in term of classification. As it uses the same OC-SVM method as the actual SVM main routine of our program, it is very efficient at grouping variables that have similar behaviors, and selecting variables that will be meaningful for the model build. One particularly interesting feature of the RFE algorithm, is that if you stop the algorithm after a few iterations, and plot the variables according to the order defined by their support vector's norms, they cluster in groups of similar variables from the most meaningful to the least meaningful one. 
\begin{figure}[H]
\includegraphics[scale=0.52]{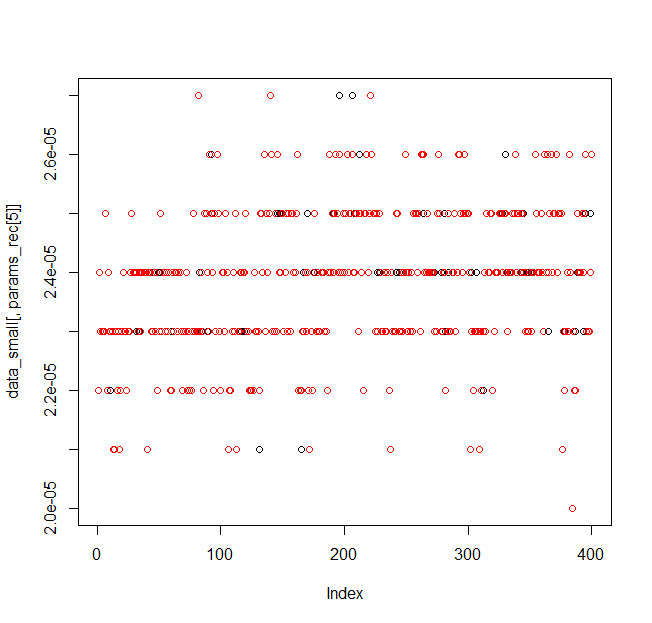}
\caption{Examples of sorted features : two variables with following ranking criterion (in black and red) are almost undistinguishable.}
\label{classif}
\end{figure}
The example in figure \ref{classif} is quite trivial, but visualization quickly becomes impossible on more realistic data sets. This behavior was exhibited and validated on 10 different training sets, with various random arrangements of the variables observations. It is quite easy to see that this is in fact a natural property of the SVM algorithm which is a classifier algorithm. It is however surprising that through the support vector norms, we exhibit a natural mapping from $\mathbb{R}^N$ to $\mathbb{R}$ of the distributions behavior, in a similar fashion as with the entropy metric presented earlier. \par
This classification property pushed us to seek for potentially optimized approaches, using K-Medoid clustering to drastically improve on the classical RFE approach in terms of both computational efficiency and performance.

\section{RFE and K-Medoid clustering.}

As mentionned earlier, after a few iterations, RFE induces a classification of the variables, grouping them according to their general form. Furthermore, the main goal of feature elimination is to select only the relevant variables, containing as much complementary information as possible. We would then want to "pick" one representative per equivalence class, the latter being defined by their ranking criterion. This can be done using K-Medoid clustering. \par
We subsequently devised the following algorithm :
\begin{itemize}
\item{Step 0 : We initiate a set of $n$ variables $\mathcal{S}=\lbrace (x_{i}^{(k)})_{i\in [0,N]} | k \in [0,n] \rbrace$ ($N$ is the number of observations of each variable).}
\item{Step 1 : Run $p$ iterations of the RFE loop until variable ranking is stable.}
\item{Step 2 : Run  a K-medoid clustering algorithm to determine the best representatives of each class of variables. The number of medoids corresponds to the final desired number of selected variables.}
\end{itemize}
This process allows to select the most representative features of a distribution, as well as selecting the most useful features from the SVM standpoint thanks to RFE. One limitation of this process, is that no extensive study of the ranking stability of RFE has been conducted. That is, experimentaly, if we follow the recursive elimination one step at a time, we can see by plotting the data that in the first few operations, variables group themselves, then the number of variables for each groups is reduced, as well as the least meaningful groups containing only a small number of variables. It would be interesting to study this clustering phenomenom more in detail to be able to devise more efficient approaches, tailored to our practical needs : defect detection. \par
We tested this method on the various labeled data sets we had at our disposal, and it prooved more effective than the previous univariate approaches, but still less than RFE. On the computation time however, it is orders of magnitude faster than RFE as K-Medoid algorithms are well-known and already heavily optimized. We used a R library based on \cite{park}. 

\section{Numerical results and performance tests.}
The actual performance of the methods described above was evaluated on two main use cases where the data was labeled. We tested different combinations of the methods :
\begin{itemize}
	\item Using only one variable selection method (either MADe, entropy or RFE).
	\item Using a hybrid method taking the intersection of the entropy and MADe methods.
	\item Using all univariate methods (for computational efficiency) and combining the results by discarding wafers that are discarded by all methods.
\end{itemize}
The results of the first use case and second use case can be found in figures \ref{usecase1} and \ref{usecase2}.
\begin{figure}[H]
	\begin{tabular}{|c|c|c|c|c|}
  	\hline
   	& MADe & Entropy & Both & Combined\\
  	\hline
  	Total & 145/1725 & 477/1725 & 196/1725 & 233/1725\\
  	ECC  & 4/13 & 8/13 & 3/13 & 10/13\\
  	\hline
	\end{tabular}
\caption{First use case}
\label{usecase1}
\end{figure}

\begin{figure}[H]
	\begin{tabular}{|c|c|c|c|c|}
  	\hline
   	& MADe & Entropy & Both & Combined\\
  	\hline
  	Total & 497/787 & 464/787 & 497/787 & 203/787\\
  	ECC  & 6/11 & 7/11 & 3/11 & 11/11\\
 	 \hline
	\end{tabular}
\caption{Second use case}
\label{usecase2}
\end{figure}

The "Total" line states the total number of wafers detected by each method. The "ECC" line states the number of actually faulty wafers detected. We see already a quite frank advantage for the combined method that strives on the different standpoints given by each method to provide the most accurate picture of the wafer possible. This favors much more efficiently the use of complementary information rather than redundant information, which was our initial goal. \par
We also see that the "both" method is not very efficient by itself as it tends to destroy too much information by taking the intersection of two variable sets and so should not be considered as is. 

\section{Estimating feature selection influence on the SVM model.}
\subsection{Different types of error.}
One major flaw of the SVM classification, especially OC-SVM is that it is very hard to estimate an error bar on the classification result. It would be greatly beneficial to be able to put an error bar on the scoring result given by the SVM, even more so on critical applications such as ours. This is why we tried to estimate different sources of error, and feature selection appeared as the prevalent one. We identified three major sources of error :
\begin{itemize}
\item{Measure error on the data fed to the SVM model (and thus its influence on the final classification result).}
\item{Feature selection error : if different variables are taken into account to build a model, then the resulting model will give slightly different results during the testing phase according to these different "standpoints".}
\item{Computation errors resulting from the resolution of a quadratic optimization problem while building the SVM model.}
\end{itemize}
The computation error can be quickly discarded as the algorithms solving the variational formulation of the SVM are very efficient and often the problem can be solved linearly using Lax-Milgram theorem. \par
The data measure error does not significantly influence the results for various reasons. Firstly, a very simple linear stability analysis shows that if measure error is small, the scoring is thus small too. Let us look at the decision $f$ function of a variable $x$ as presented in \cite{Lampert}. Let $(x_i)_{i\in [1,n]}$ be the learning sample, then for any new point $x$ :
\begin{equation}
f(x)=\sum_{i=1}^{n}\alpha_i K(x,x_i) - \rho
\end{equation}
If we move the learning sample by a small amount $\delta$ that is $x_i=x_i + \delta_i$, assuming that the coefficients $\alpha_i$ remain constant, we get :
\begin{equation}
\delta f(x)=\sum_{i=1}^{n}\alpha_i \delta_i \frac{\partial K}{\partial x_i}(x,x_i) 
\end{equation}
Experimentally, $\frac{\partial K}{\partial x_i}(x,x_i) $ is very small, and if we introduce a gaussian noise vector as $\delta$, the differences tend to balance out so we can reasonably assume that measure noise on the learning sample has no effect. \par
The only source of error left is then feature selection as we litterally swap out some $x_i$ terms in the sum. Having no prior knowledge of the features to be selected, an experimental approach is to be prefered.

\subsection{Experimental procedure for feature selection error bars.}
As mentioned before, when building a model using feature selection, we have no prior knowledge of what feature is going to be selected, especially with RFE. We devised an experimental scheme on random data to explore the effect of feature selection. And we foud a correlation between the variations of two quantities, a correlation that can later be used to estimate error bars while testing the model on another sample. \par
We generated random data using a uniform distribution, and a normal distribution. We then built a hundred models for each data set (taking the first half of the set as a learning sample), randomly removing ten variables for each model to emulate variable selection. We then tested the models on the other half of the test.  \par
We found out that there was a linear relationship between the standard deviation of the scores of one sample, and the standard deviation of a single variable over all samples. This is very interesting because, we can only access to the first standard deviation while working on actual datasets. Furthermore, this ratio was exprementally proved to be a constant only depending on the fraction of variables removed during feature selection. This means that during the training phase, we can compute both standard deviations and thus their ratios, and finally predict the standard deviation of the scoring over a single test sample during the testing phase. \par
On actual data sets, we found that this standard deviation discarded the 1\% of variables closest to the limit. This "grey zone" is especially meaningful for industrial applications where the number of lots being tested are very big. 

\section{Conclusion and outlook}

The results obtained were very promising and allowed a lot of progress in the field of quality control at ST, where roughly 50\% of all of the produced wafers get stopped at some point of their production for further analysis before continuing production. We then see that the presented methods would reduce this rate to a mere 13.5\% as seen in figure \ref{usecase1}. Moreover compared to the previous methods used for quality control, the SVM approach allows to distinguish patterns on specific production parameters that can then be used as guidelines for production engineers to correct for drifts in the various processes used in the fabrication process. The next step would be to implement the same kind of monitoring using SVM on the machine themselves to track their parameters and then detect cross-correlations. 
However, on a purely scientifical point of view, a lot of questions remain unanswered and would need further research. Firstly, the purely mathematical aspect of feature selection is not yet very clearly understood. It is fairly easy to grasp why some methods yield better results than other, but no formal proof was studied (mainly due to a limited time). Secondly, the two most integrated approaches to feature selection, namely the RFE and K-med methods that were presented, did not show the expected results, especially considering what can be found in the litterature \cite{Huang}. More thoughts need to be put in efficient implementations of RFE, as well as a work on K-med clustering as a way to speed up RFE, but this is very tedious considering the computation times needed to test our theories. Finaly, it would be very interesting to study the influence of kernel choice for our industrial data, especially kernels with exotic distances such as Mahalanobis distance, as we can suspect it could have a great influence on the SVM's performance by introducing a multivariate criterion from the start.

\section*{Acknowledgements}
We would like to thank the whole quality control team of STMicroelectronics Crolles for their fruitful suggestions and practical remarks.
\end{multicols}

\newpage
\bibliography{bib_article}

\end{document}